\documentclass[sigconf,nonacm]{acmart}

\AtBeginDocument{%
  }

\setcopyright{none}
\renewcommand\footnotetextcopyrightpermission[1]{}

\acmConference[ACM MM '26]{Proceedings of the 34th ACM International Conference on Multimedia}{October 2026}{Dublin, Ireland}
\acmBooktitle{Proceedings of the 34th ACM International Conference on Multimedia (ACM MM '26), October 2026, Dublin, Ireland}
\acmDOI{XXXXXXX.XXXXXXX}
\acmISBN{978-1-4503-XXXX-X/26/10}

\usepackage{booktabs}

\title{Look, Zoom, Understand: The Robotic Eyeball for Embodied Perception}

\author{Jiashu Yang}
\affiliation{%
  \institution{School of Artificial Intelligence, Shanghai Jiao Tong University}
  \city{Shanghai}
  \country{China}
}
\email{jiashuyang827@gmail.com}

\author{Yifan Han}
\affiliation{%
  \institution{Institute of Automation, Chinese Academy of Sciences}
  \city{Beijing}
  \country{China}
}

\author{Yucheng Xie}
\affiliation{%
  \institution{Dalian University of Technology}
  \city{Dalian}
  \country{China}
}

\author{Ning Guo}
\affiliation{%
  \institution{School of Artificial Intelligence, Shanghai Jiao Tong University}
  \city{Shanghai}
  \country{China}
}

\author{Wenzhao Lian}
\authornote{Corresponding author}
\affiliation{%
  \institution{School of Artificial Intelligence, Shanghai Jiao Tong University}
  \city{Shanghai}
  \country{China}
}

\begin{abstract}
In embodied AI, visual perception should be active rather than passive: the system must decide where to look and at what scale to sense to acquire maximally informative data under pixel and spatial budget constraints. Existing vision models coupled with fixed RGB-D cameras fundamentally fail to reconcile wide-area coverage with fine-grained detail acquisition, severely limiting their efficacy in open-world robotic applications. We study the task of \textit{language-guided active visual perception}: given a single RGB image and a natural language instruction, the agent must output pan, tilt, and zoom adjustments of a real PTZ (pan-tilt-zoom) camera to acquire the most informative view for the specified task. We propose \textbf{EyeVLA}, a unified framework that addresses this task by integrating visual perception, language understanding, and physical camera control within a single autoregressive vision-language-action model. EyeVLA introduces a semantically rich and efficient hierarchical action encoding that compactly tokenizes continuous camera adjustments and embeds them into the VLM vocabulary for joint multimodal reasoning. Through a data-efficient pipeline comprising pseudo-label generation, iterative IoU-controlled data refinement, and reinforcement learning with Group Relative Policy Optimization (GRPO), we transfer the open-world understanding of a pre-trained VLM to an embodied active perception policy using only 500 real-world samples. Evaluations on 50 diverse real-world scenes across five independent evaluation runs demonstrate that EyeVLA achieves an average task completion rate of 96\%. Our work establishes a new paradigm for instruction-driven active visual information acquisition in multimodal embodied systems.
\end{abstract}



\begin{document}

\pagestyle{plain}

\maketitle

\section{Introduction}

\begin{figure}[t]
    \centering
    \includegraphics[width=\columnwidth]{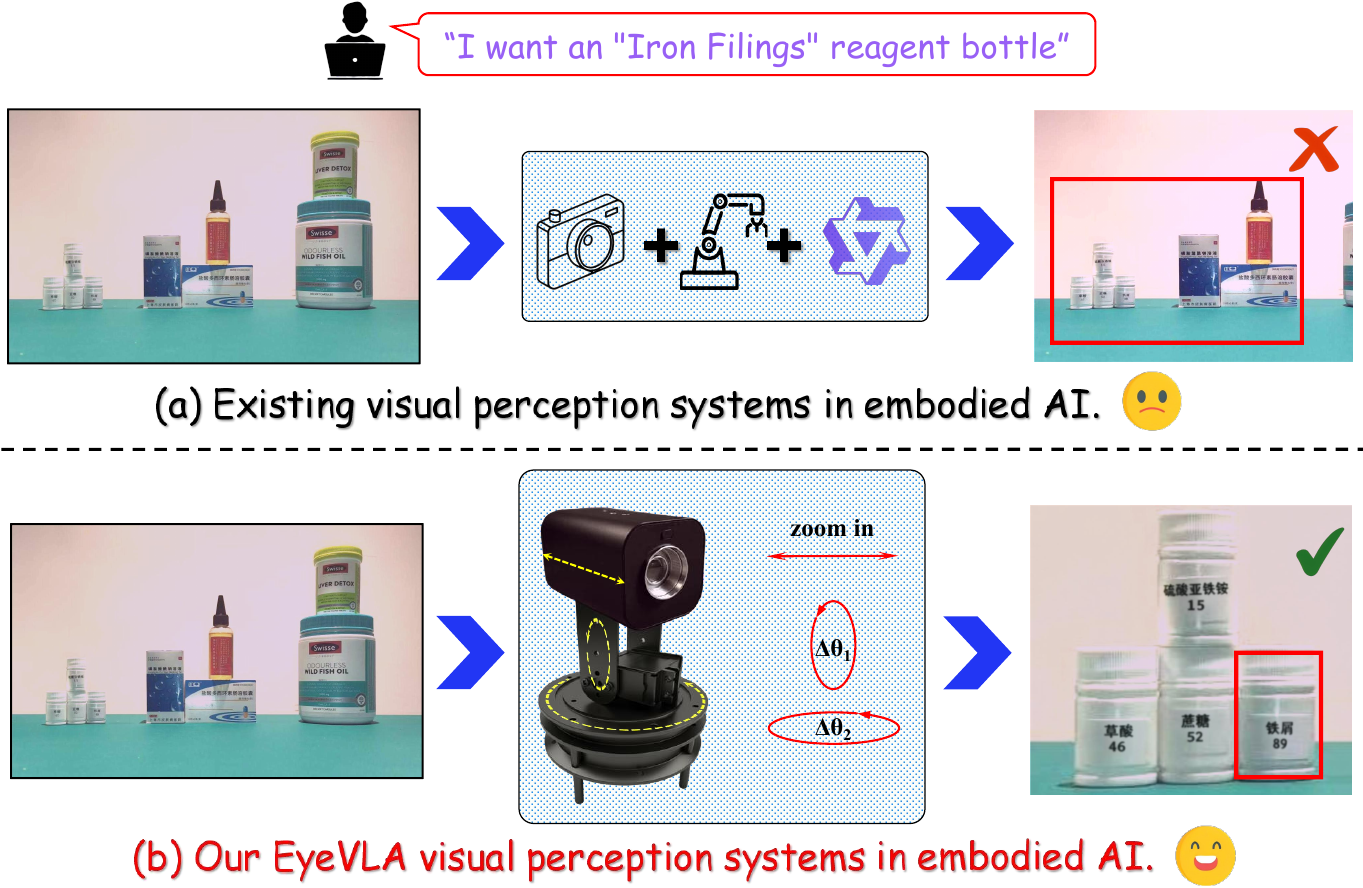}
    \caption{\textbf{(a)}: Existing vision systems with fixed RGB-D cameras cannot handle fine-grained visual information across larger spatial extents. \textbf{(b)}: Our EyeVLA system can perceive broader and finer-grained visual information from a fixed position by rotating its viewpoint and zooming in on the target, according to instructions.}
    \label{fig:motivation}
\end{figure}

In recent years, Vision-Language Models (VLMs)~\cite{liu2023visual,hurst2024gpt4o,team2023gemini,bai2025qwen25vltechnicalreport,alayrac2022flamingo} have dramatically advanced zero-shot recognition, instruction following, visual question answering, and multimodal reasoning. Modern VLMs already ``understand what is in an image'' to a degree that shifts the primary bottleneck in many embodied settings away from semantic inference quality towards the upstream process of acquiring the right visual evidence. Crucially, nearly all prevailing VLM paradigms still assume passively supplied, information-sufficient, static images as input. They operate as passive observers, lacking the agency to decide where to look, at what scale to sense, or how to adaptively reallocate limited pixel, time, and sensing resources to resolve ambiguity.

In real-world embodied scenarios, however, perception is rarely a one-shot encoding problem; it is a sequential, resource-constrained decision process. Robots frequently need to perceive distant, fine-grained details—such as reading small text on a medicine bottle, identifying a specific tool connector, or verifying the state of a tiny toggle switch—to make informed decisions. Relying on physical locomotion to approach these targets is highly inefficient in terms of time, energy, and path planning overhead. Current systems attempt to meet these demands with fixed sensors. Yet fixed monocular or RGB-D cameras—even when coupled with powerful VLM backbones—inherently interleave information redundancy (large low-value background areas) with critical detail omissions. They cannot resolve distant text or small task-decisive structures without sacrificing wide-area coverage. As a result, conventional VLM-based pipelines underperform on precision-sensitive tasks, despite excelling at global scene summarization.

The gap is clear: the semantic understanding of VLMs is already strong, yet the ability to actively acquire the right visual evidence remains missing. Furthermore, while recent Vision-Language-Action (VLA) models~\cite{brohan2023rt, kim2024openvla} have made strides in embodied AI, they primarily focus on \textit{robotic manipulation}—predicting end-effector or joint-level motor commands to interact with objects. In contrast, we argue that before a robot can manipulate the world, it must first master \textit{perceptual action}. Instead of moving the entire robot body, a more elegant and efficient solution is to move the ``eyes.''

To this end, we define the task of \textit{language-guided active visual perception}: given a single RGB image captured by a PTZ (pan-tilt-zoom) camera and a natural language instruction describing the information need (e.g., ``What is the brand of the pen?''), the system must predict three values—pan ($\Delta\theta_1$), tilt ($\Delta\theta_2$), and zoom ($\Delta z$)—to reposition and rescale the field of view so that the task-relevant visual detail is centered and magnified. Unlike manipulation, our action space directly controls the observation process on real PTZ actuators.

To address this task, we propose \textbf{EyeVLA}, a language-guided active vision framework that unifies vision, language, and embodied camera control within a single autoregressive model. By embedding action tokens directly into the VLM vocabulary, we seamlessly transform a pre-trained VLM into a VLA agent without requiring a separate control head. A core challenge in this transformation is representing continuous PTZ values. Naive continuous regression often struggles to converge within discrete language modeling architectures. To solve this, we introduce a hierarchical action encoding that discretizes continuous camera adjustments into structured tokens. This design is not only highly token-efficient but also preserves strong semantic priors, making it significantly easier for the model to learn the mapping from visual intent to physical camera movements.

Training such a VLA model requires action-annotated data, but manually collecting real-world PTZ demonstrations is prohibitively expensive and difficult to scale. We overcome this bottleneck through a data-efficient post-training pipeline. We first synthesize pseudo-data from existing large-scale grounding datasets by mapping bounding box coordinates to simulated PTZ actions. However, directly mapping from an image to camera actions presents a steep learning curve. Therefore, during training, we incorporate 2D bounding-box predictions as chain-of-thought-style (CoT) spatial reasoning cues, guiding the model to explicitly localize the target region before generating actions. We further employ iterative IoU-based filtering to refine the quality of the pseudo-labels. Finally, because synthetic data cannot perfectly capture real-world hardware tolerances and scene-dependent geometry, we apply reinforcement learning (RL) to align the policy with real-world action distributions and enhance robustness.

Extensive ablations and evaluations across 50 diverse real-world scenes over five independent evaluation runs demonstrate the effectiveness of our approach, achieving an average task completion rate of 96\%. In summary, our contributions are as follows:

(i) We formulate language-guided active visual perception as a discrete, tokenized decision process integrated with multimodal reasoning, transforming visual perception from passive frame consumption into closed-loop, task-aware active acquisition.

(ii) We introduce a semantically rich and token-efficient hierarchical action encoding that embeds pan/tilt/zoom adjustments into the VLM vocabulary, enabling unified autoregressive modeling without a separate control head.

(iii) We design a data-efficient training pipeline that leverages pseudo-data generation, CoT-style bounding-box guidance, and iterative IoU filtering, followed by RL to align the policy with real-world distributions and enhance robustness.

(iv) We demonstrate that with only 500 real-world samples and pseudo-labeled expansions, EyeVLA achieves an average task completion rate of 96\% across 50 diverse real-world scenes over five independent runs, establishing a practical path toward low-cost embodied active perception.

\section{Related Work}
\noindent \textbf{Vision Language Models.} Recent advances in vision language models have enabled open-vocabulary grounding and instruction-following reasoning. Early work such as CLIP \cite{radford2021learning} aligns visual and textual representations for zero-shot image classification but lacks fine-grained spatial localization. GLIP \cite{li2022grounded} reformulates object detection as phrase grounding, enabling zero-shot localization via natural language queries. Grounding DINO \cite{ren2024groundingdino15} integrates DETR with a language encoder for high-precision, end-to-end open-vocabulary detection. In parallel, generative VLMs have advanced multimodal instruction following. LLaVA~\cite{liu2023visual} and InstructBLIP~\cite{dai2023instructblip} bridge the visual encoder and LLM via a projector, achieving alignment between vision and language. Further, Qwen-VL~\cite{bai2023qwen,bai2025qwen25vltechnicalreport} and InternVL~\cite{chen2024internvl} series optimized visual token compression and cross-modal alignment, achieving breakthroughs in multi-image reasoning, long-context localization, and instruction generalization, demonstrating strong general-purpose capabilities. Despite the progress in semantic and spatial understanding, existing VLMs operate on static, passively captured images, lacking control over the visual acquisition process. Our work addresses this by enabling language-guided active visual perception in dynamic environments.

\noindent \textbf{Vision Language Action Models.} VLA models extend VLMs to embodied control by mapping multimodal inputs directly to executable actions. Early approaches embed VLM features into policy architectures or visuomotor pipelines~\cite{shridhar2022cliport,stone2023open}, but often depend on hand-crafted structures or calibrated cameras, limiting scalability in real-world robotics. More recent methods instead fine-tune large pretrained VLMs for action generation, leveraging web-scale vision--language pretraining and general-purpose transformer backbones to align perception and control at scale~\cite{brohan2023rt,huang2023embodied,li2023vision,kim2024openvla}. However, existing VLA models typically either attach a simple action head on top of VLM representations or directly represent actions as discrete text tokens~\cite{liu2024robomamba,wen2025tinyvla,yang2025efficientvla,goyal2025vla}, which introduces an information bottleneck and makes it difficult to learn high-precision, fine-grained numerical control. Crucially, all the above VLA works target robotic manipulation—predicting end-effector or joint-level motor commands to interact with objects. In contrast, our work addresses a fundamentally different problem: active visual perception, where the action space consists of camera pan, tilt, and zoom adjustments on real PTZ actuators. The objective shifts from ``how to act on the world'' to ``how to acquire the most informative observation for downstream understanding.'' This distinction necessitates a different action representation (our hierarchical encoding) and training paradigm (bounding-box-guided RL for information maximization rather than manipulation success).

\begin{figure*}[ht]
    \centering
    \includegraphics[width=\textwidth]{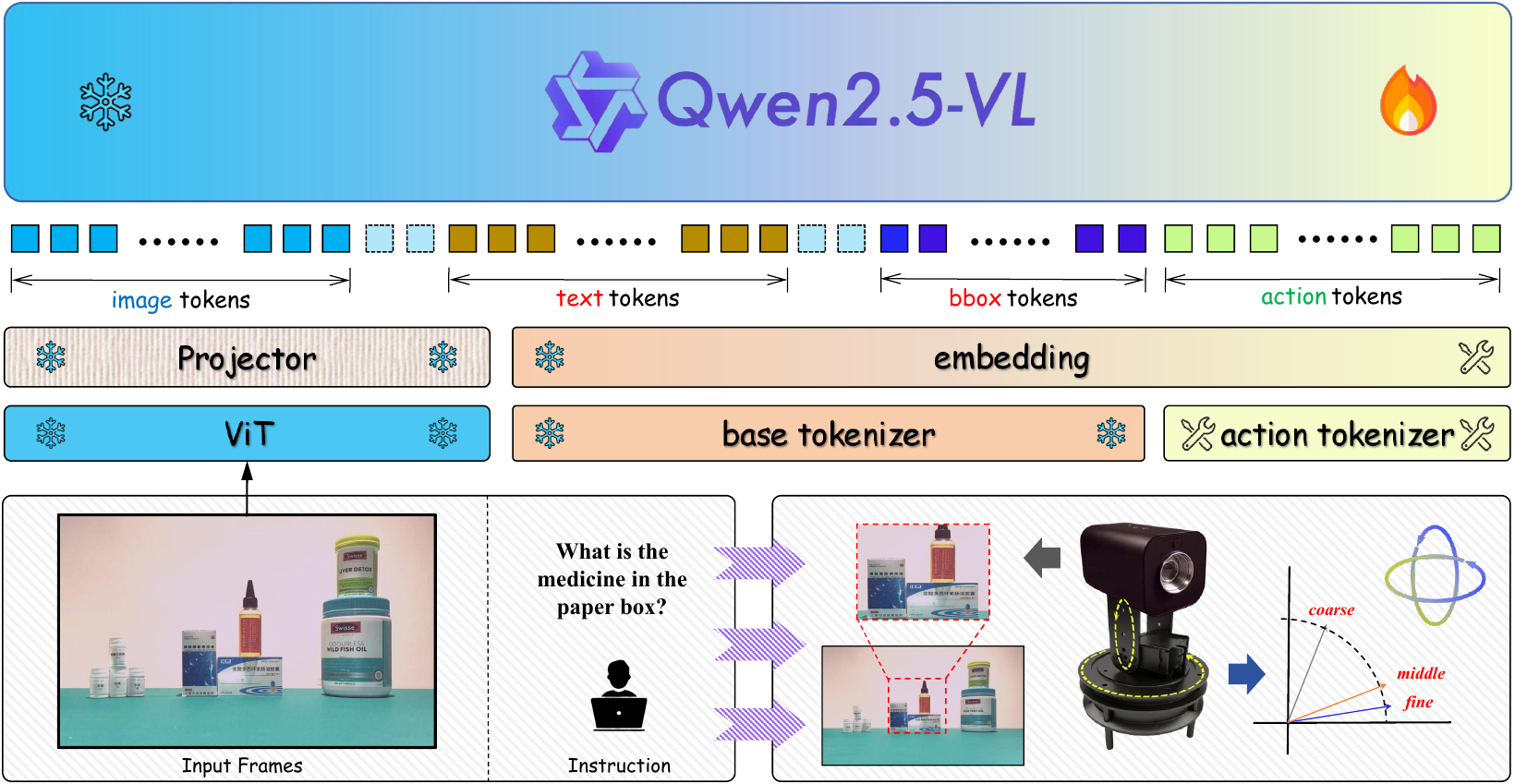}
    %\vspace{+0.1in}
    \caption{\textbf{Overview of EyeVLA Pipeline.} The system is built upon  Qwen2.5-VL framework, integrating visual perception, language understanding, and action generation capabilities. To preserve the original semantic alignment during training, the parameters of the ViT and its projector module are kept frozen and not updated. Additionally, we introduce action tokens into the vocabulary to represent camera motions. To efficiently represent robotic actions, we further adopt a hierarchical encoding strategy to structurally model the action space.}
    \label{fig:pipeline}
\end{figure*}

\noindent \textbf{Active Perception in Robotics.} Recently, active perception has increasingly been explored in conjunction with vision-language models, enabling robots to actively acquire task-relevant visual information rather than relying on static observations. To support such closed-loop perception and decision-making, specialized robotic hardware architectures with active camera mobility and controllable viewpoints have been developed~\cite{xiong2025via, chuang2025active, liu2025avractivevisiondrivenprecise}.
By constructing a VLM-based probabilistic sensor model and incorporating semantic uncertainty into geometric-semantic map representations, prior work enables robots to actively select observation viewpoints through uncertainty-driven exploration policies~\cite{bajpai2025uncertainty}. Within this paradigm, Sripada et al.~\cite{sripada2025sceneexplorationvisionlanguagemodels} adopt language-guided discrete Next-Best-View on predefined grids but lack continuous control. Conversely, VG-AVS~\cite{koo2025toward} applies reinforcement learning to NBV for visual question answering, while MTU3D~\cite{zhu2025move} enables end-to-end trajectory learning for object grounding without explicit 3D reconstruction. Similarly, Active-O3~\cite{zhu2025active} extends VLMs with semantic-driven viewpoint selection at the level of software-based algorithmic design, where viewpoint search is abstracted without physical embodiment or real camera actuation. Although many existing systems demonstrate the value of language-guided active perception, they often rely on predefined action spaces and cascaded modules, prone to accumulated modular errors and lacking tight integration between visual/language reasoning and low-level motor control. In contrast, our system unifies language grounding, action generation, and visual feedback within a single autoregressive loop, enabling closed-loop active perception under pixel and spatial budget constraints.

\section{Method}\label{sec:method}

\subsection{Overall Approach}

In this section, we elaborate on the proposed \textbf{EyeVLA} framework. The system consists of a hardware platform, the \textit{Robotic EyeBall} (REB) comprising a two-axis pan-tilt mount and a zoomable camera, and an algorithmic pipeline that adapts Qwen2.5-VL (7B) for joint perception and camera control (see Figure~\ref{fig:pipeline}). At inference time, the system receives an initial wide-field image together with a natural language instruction, predicts a compact action triplet $\mathbf{a}$, and drives the physical actuators to acquire a zoomed-in, task-relevant view in a single forward pass, without iterative search or manual intervention.

\noindent \textbf{Problem Formulation.}
We formally define the task of \textit{language-guided active visual perception}. Let $\mathbf{I}_0 \in \mathbb{R}^{H \times W \times 3}$ denote an initial RGB image captured by a PTZ camera at its default pose, and let $\mathcal{L}$ denote a natural language instruction specifying the information need (e.g., ``What is the brand of the pen?''). The objective is to learn a policy $\pi$ that predicts a camera adjustment action $\mathbf{a} = (\Delta\theta_1, \Delta\theta_2, \Delta z)$, where:
\begin{itemize}
    \item $\Delta\theta_1 \in \mathbb{R}$: horizontal pan angle change;
    \item $\Delta\theta_2 \in \mathbb{R}$: vertical tilt angle change;
    \item $\Delta z \in \mathbb{R}_{>0}$: zoom magnitude change.
\end{itemize}
The predicted action drives the PTZ actuator to reposition and rescale the camera's field of view, yielding a new observation $\mathbf{I}_1$ in which the task-relevant visual detail is centered and maximally resolved. Formally, we seek:
\begin{equation}
    \mathbf{a}^{*} = \arg\max_{\mathbf{a}} \; \mathcal{R}(\mathbf{I}_1(\mathbf{a}), \mathcal{L}),
\end{equation}
where $\mathcal{R}$ measures the task-relevant information gain in the resulting view $\mathbf{I}_1(\mathbf{a})$ with respect to the instruction $\mathcal{L}$. In practice, we instantiate $\pi$ as an autoregressive VLM that generates a sequence of discrete action tokens, which are then decoded into the continuous action triplet $\mathbf{a}$. By autoregressively predicting these tokens, the system drives the Robotic EyeBall to autonomously adjust its viewpoint in 3D space, aligning the resulting visual state with the semantic intent expressed in the instruction.

\begin{figure*}[t]
    \centering
    \includegraphics[width=\textwidth]{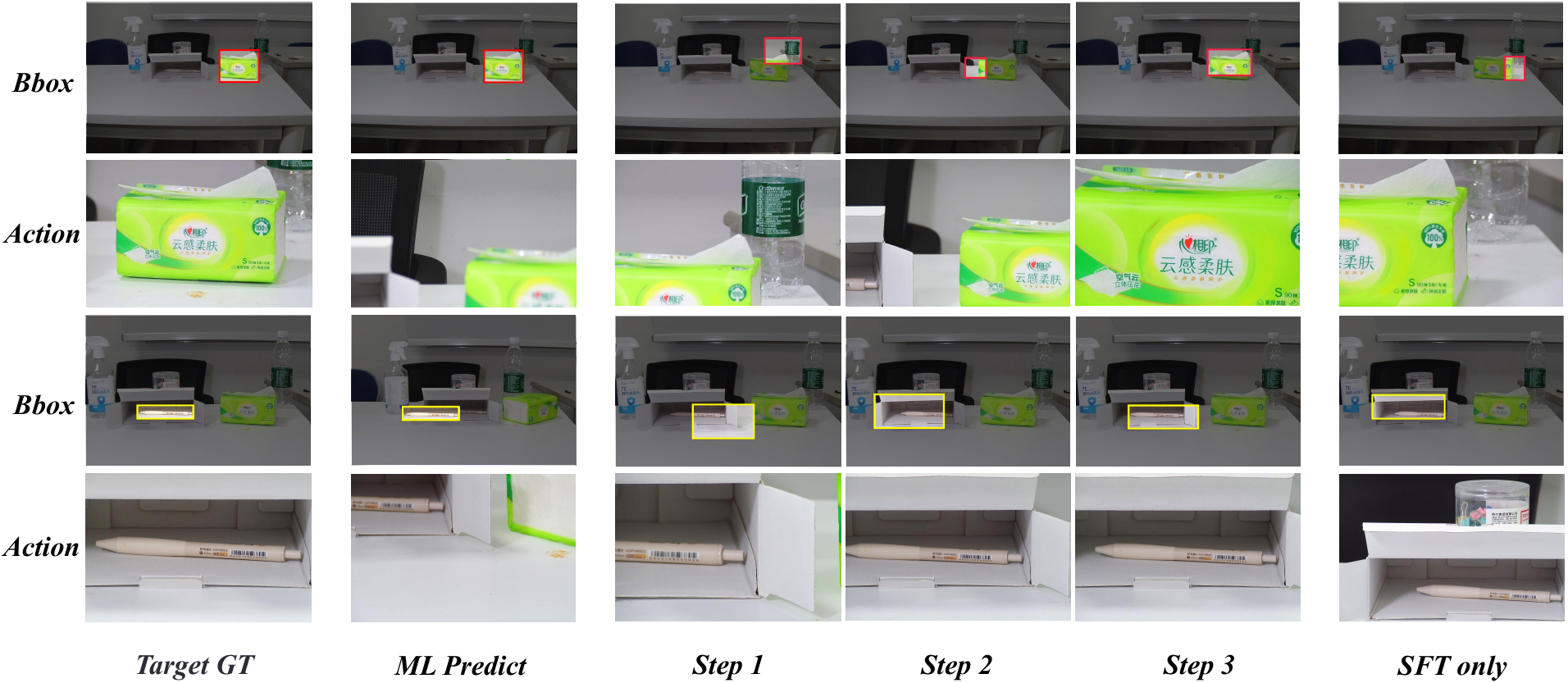}
    \caption{Inference results of models trained on synthetic data generated under different strategies and iteration counts, along with their performance comparison in real-world scenarios. The figure illustrates a scenario where the goal is to identify the brand of a pen inside a box. A conventional camera, constrained by its fixed viewpoint, cannot extend into the box via a robotic arm to capture fine details. In contrast, our EyeVLA system enables clear view of the target by dynamically adjusting the camera pose and zooming in. }
    \label{fig:3}
\end{figure*}

\subsection{Hierarchical Action Encoding}

To enable a compact and token-efficient representation of continuous camera actions within a limited context length, we propose a hierarchical action encoding scheme that minimizes the number of tokens required to represent integer-valued pan and tilt angle changes ($\Delta\theta_1, \Delta\theta_2$) as well as zoom adjustments ($\Delta z$), while maintaining semantic coherence with the original text tokens.

Specifically, for a target angle $x \in \mathbb{Z}_{\geq 0}$, we first decompose it into decimal digits:
\begin{equation}
x = \sum_{\ell=0}^{L-1} x^{(\ell)} \cdot 10^{\ell}, \quad x^{(\ell)} \in \{0, 1, \dots, 9\},
\end{equation}
where $\ell$ denotes the digit level ($\ell=0$: units, $\ell=1$: tens, etc.), and $L$ is the maximum number of digits (empirically, $L=2$ suffices for over 98\% of real-world actions).

For each digit $x^{(\ell)}$, we further encode it as a non-negative integer linear combination of the basis set $D = \{5, 2, 1\}$:
\begin{equation}
x^{(\ell)} = \sum_{d \in D} a_d^{(\ell)} \cdot d, \quad a_d^{(\ell)} \in \mathbb{Z}_{\geq 0},
\end{equation}
where $a_d^{(\ell)}$ represents the coefficient for each basis element $d \in D$. The encoding objective is to minimize the token count per digit:
\begin{equation}
\min_{\{a_d^{(\ell)}\}} \sum_{d \in D} a_d^{(\ell)} \quad \text{subject to} \quad \sum_{d \in D} a_d^{(\ell)} \cdot d = x^{(\ell)}.
\end{equation}
This formulation corresponds to the classic \textit{Change-Making Problem} restricted to the domain $\{0, \dots, 9\}$.

Critically, the basis $D = \{5, 2, 1\}$ constitutes a \textit{canonical coin system}. By the result of Kozen and Zaks~\cite{kozen1994optimal}, the greedy algorithm (iteratively selecting the largest denomination not exceeding the remaining value) yields a globally optimal solution (i.e., minimal token count) for every $x^{(\ell)} \in [0, 9]$. We encode each digit place (i.e., each power of 10) using this efficient representation.

Consequently, during decoding, the original angle can be reconstructed by summing the contributions of each basis element per digit level. The greedy strategy ensures that both encoding and decoding are computationally efficient with $O(1)$ per-digit complexity.

This design offers three key advantages: (1) theoretical optimality in token usage within each digit; (2) constant-time encoding and decoding; and (3) structural alignment with physical camera control, where higher digits correspond to coarse adjustments and lower digits to fine tuning. Empirical analysis on 500 real-world samples shows that 98.6\% of actions fall within $\pm 29^\circ$, requiring at most two digit levels and achieving an average token length of 2.3, which is significantly more efficient than uniform discretization (one token per degree, yielding an average length of 12.7).

\subsection{Data Generation and Iterative Refinement}

\noindent \textbf{Initial Data Collection.} We collect initial demonstration data by manually operating the robotic eye system to align with human-specified natural language instruction targets. Given an initial frame captured by the camera at its default pose, a human operator issues a goal-oriented instruction (e.g., ``What is the brand of the pen?''). The operator then manually adjusts the 2D pan-tilt mount and zoomable camera to center and magnify the target object, recording the corresponding change signals:
\begin{align*}
   \Delta\theta_1 & : \text{horizontal pan displacement} \\
   \Delta\theta_2 & : \text{vertical tilt displacement} \\
 \Delta z & : \text{zoom step change}
\end{align*}
This process yields a dataset of (instruction, initial frame, $\Delta\theta_1$, $\Delta\theta_2$, $\Delta z$, post-action frame) tuples, where the action triplet precisely realizes the intent expressed in the command. To promote diversity, demonstrations were collected across multiple indoor environments featuring objects of varying sizes, distances, and semantic categories, ensuring that the dataset captures a wide range of spatial configurations.

\noindent \textbf{Pseudo-Label Engine.} Due to the extremely high time and labor costs of manually collecting real-world data, and to enable the model to possess strong zero-shot capabilities in open-world scenarios, we synthesize pseudo-labeled data whose distribution approximates the real demonstrations. For low-cost task transfer, we utilize existing grounding datasets (including images, referring expressions, and annotated bounding boxes) for data synthesis.

Specifically, we normalize the center coordinates of the bounding box (bbox) from the initial image into the range $(-1, 1)$ to model the horizontal (pan) and vertical (tilt) angle relationships. For zoom modeling, we use the ratio of the bbox area relative to the total image area. Formally, let $w_1 = \frac{s_1}{s}$ denote the ratio of the bbox area $s_1$ to the total image area $s$ in the pre-zoom image, and $w_2 = \frac{s_2}{s}$ denote the corresponding ratio in the post-zoom image. We model the zoom action using the ratio $w_1 / w_2$.

From the grounding dataset, we select 50,000 samples with the smallest target area ratios (prioritizing small, distant, or partially occluded targets), because such configurations most closely resemble the practical scenarios in which PTZ active perception is needed. Using the bbox descriptions, we construct textual prompts and apply a Random Forest model (trained on our real-world demonstrations) to generate pseudo-labels. Since our objective is to place the target object as large and centered as possible, we perform isotropic cropping to maximize the bbox area. The cropped image serves as the adjusted ground truth, based on which we recompute the area ratio to obtain an updated $w_2$, subsequently generating the pseudo-labeled zoom value. Finally, based on our hierarchical vocabulary, we use a greedy algorithm to encode the continuous pan, tilt, and zoom values into discrete action tokens.

\begin{figure}[t]
\centering
\includegraphics[width=\columnwidth]{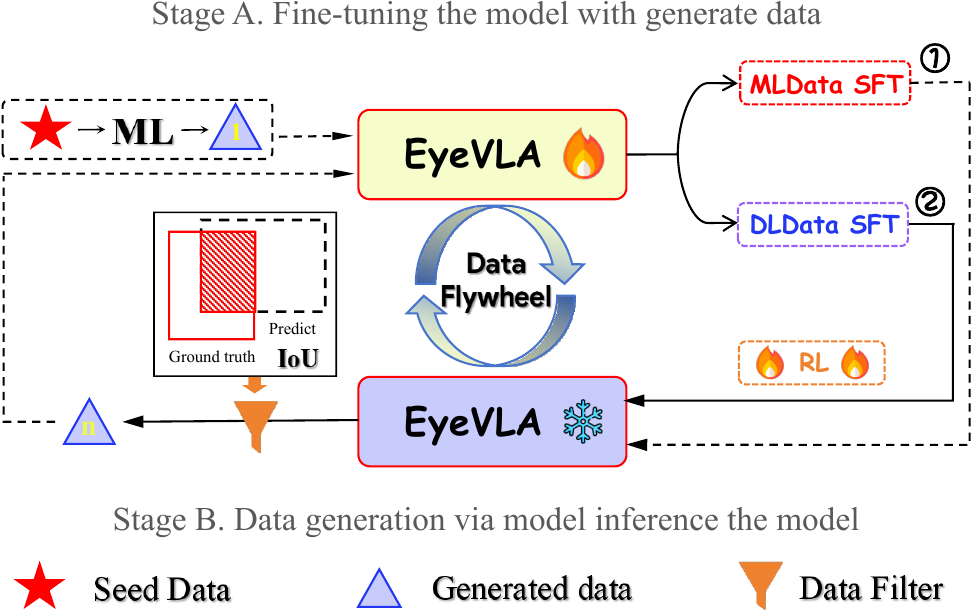} % 留点边距更安全
\caption{Flowchart of a Data Generator.}
\label{fig:4}
\end{figure}

\subsection{Training Strategy: Iterative SFT and RL}

Our training procedure consists of two main phases: Supervised Fine-Tuning (SFT) to inject action generation capabilities while preserving vision-language alignment, and Reinforcement Learning (RL) to guide the model toward real-world generalization.

\paragraph{Phase 1: Iterative Supervised Alignment with CoT Guidance}
We first perform SFT on the vocabulary-extended VLA model using the generated pseudo-labeled samples. To preserve the visual representation capabilities acquired during pre-training, we freeze the Vision Transformer (ViT) encoder and the vision-language projector. Only the language model backbone and the newly introduced action token embeddings are updated.

To bridge the gap between raw images and camera actions, we incorporate 2D bounding-box predictions as chain-of-thought-style (CoT) spatial reasoning cues. The model is trained to first predict the target's bbox before generating the action tokens. This two-step formulation forces the model to explicitly localize the target in pixel space before committing to a camera control decision, providing an interpretable intermediate representation that grounds the action prediction in visual evidence rather than mapping directly from image to action in a single opaque step.

\begin{table*}[t]
\centering
\renewcommand{\arraystretch}{1.5}
\setlength{\tabcolsep}{1.32mm}
\caption{
The table presents the MAE and completion rate (CR) of three actions under different methods and training stages: $\theta_1$ and $\theta_2$ denote rotational errors (in degrees, °), and Zoom represents the error in zoom-in magnitude (a change of approximately 90 corresponds to 1$\times$ zoom). Lower values indicate better performance. "ML" refers to the machine learning baseline; "SFT1" and "SFT2" denote the supervised fine-tuning results at the second and third stages, respectively; "SFT(Y)" indicates that IoU-based sample selection was used, while "SFT(N)" denotes its absence; "RL2" and "RL3" correspond to RL training conducted after SFT2 and SFT3, respectively.
}
\begin{tabular}{lcccccccc}
\toprule
Metric & ML & SFT1 & SFT2 (Y) & SFT2 (N) & SFT3 (Y) & SFT3 (N) & RL2 & RL3 \\
\midrule
$\theta_1$  & 6.44° & 6.39° & 3.97° & 4.91° & 2.94° & 3.99° & 3.26° & 2.04° \\
$\theta_2$  & 3.25° & 3.09° & 2.38° & 4.87° & 1.93° & 3.62° & 2.16° & 1.68° \\
Zoom  & 137.60 & 118.18 & 90.68 & 130.37 & 68.22 & 99.72 & 71.33 & 65.37\\
CR  & 36\% & 36\% & 72\% & 56\% & 92\% & 84\% & 80\% & 96\%\\
\bottomrule
\end{tabular}
\label{tab:tab1}
\end{table*}

We adopt an iterative refinement strategy for SFT:
\begin{itemize}
    \item \textbf{SFT1}: We train the base model (Qwen2.5-VL-7B-Instruct) on the initial synthetically generated samples to learn the basic representation of action tokens.
    \item \textbf{SFT2}: We use the SFT1 model to infer action tokens on the synthetic samples. We then filter these samples, keeping only those where the Intersection over Union (IoU) between the predicted and ground-truth bounding boxes is greater than 0.7. We replace the predicted bboxes with ground-truth bboxes to construct a refined dataset, and perform the second round of SFT.
    \item \textbf{SFT3}: We repeat the refinement process, raising the filtering threshold to an IoU of 0.95 to further ensure the high quality of the pseudo-labels.
\end{itemize}

\paragraph{Phase 2: Policy Refinement via Reinforcement Learning}
After the iterative SFT alignment, residual discrepancies remain between the pseudo-labels and real-world execution due to hardware tolerances and scene-dependent geometry. We therefore apply Reinforcement Learning (RL) to optimize the policy on the real-world collected data, correcting potential biases and improving robustness. We utilize Group Relative Policy Optimization (GRPO)~\cite{wei2024grpo}, which leverages group-normalized advantages and clipped probability ratios to stabilize training.

\begin{figure*}[t] % [t] 表示尽量放在页面顶部
\centering
\includegraphics[width=\textwidth]{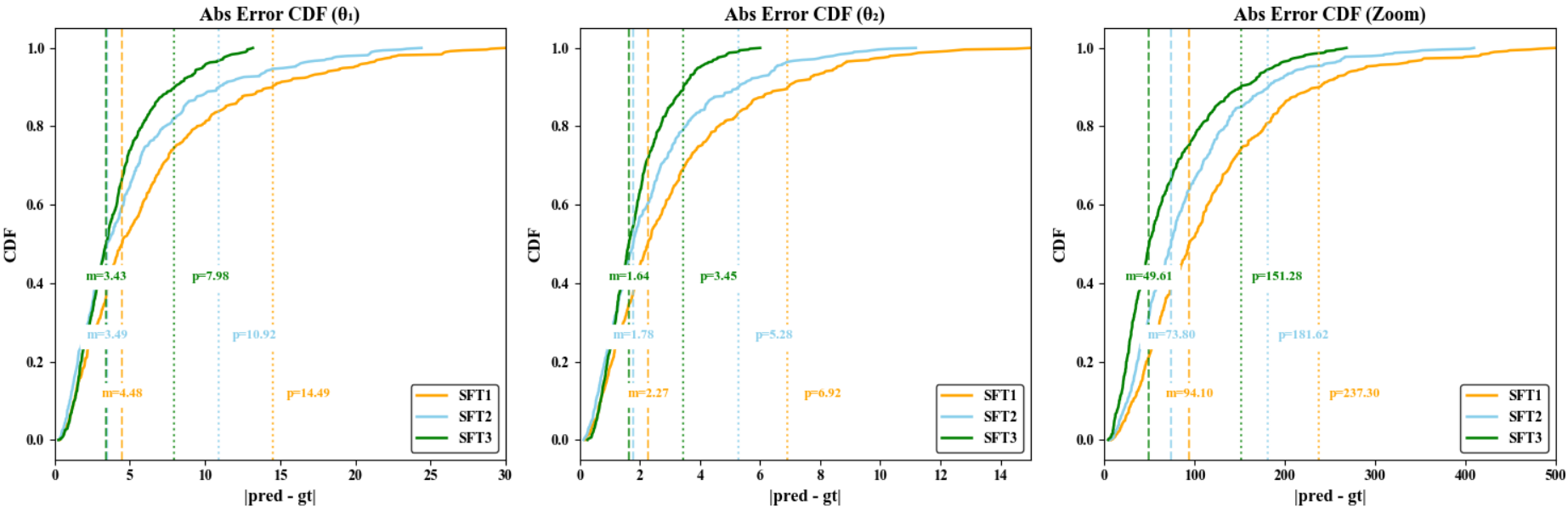}
\caption{ Comparison of Results from Three-Stage SFT}
\label{fig:011}
\end{figure*}

The GRPO objective is:
\begin{align}
\mathcal{J}(\theta) 
& = \frac{1}{N} \sum_{i=1}^{N} \Big[ 
 \min\big(s_i(\theta),\, \mathsf{clip}(s_i(\theta), 1 - \epsilon, 1 + \epsilon)\big) A_i  \nonumber \\
& {} - \beta D_{\mathsf{KL}}\!\left( 
\pi_{\theta}(\cdot \mid q) \, \| \, \pi_{\mathsf{ref}}(\cdot \mid q) 
\right) \Big] ,
\end{align}
where $\pi_{\mathsf{ref}}$ denotes the initial VLM, $\epsilon$ defines the trust region for the importance ratio $s_i(\theta) = \frac{\pi_{\theta}(o_i | q)}{\pi_{\theta_{\mathsf{old}}}(o_i | q)}$, and $\beta$ is the weight applied to the KL divergence penalty.

This iterative training pipeline effectively transfers open-world knowledge from synthetic data to physically grounded camera control, with real-world samples serving as corrective RL feedback. The resulting model jointly reasons over vision, language, and action within a single autoregressive pass.

\section{Experiment}

\subsection{Experimental Settings}

\noindent \textbf{Dataset.}
We have developed a robotic eye system comprising a two-axis pan-tilt mount and a zoomable camera. We collected a total of 500 real-world demonstration samples spanning diverse object categories and indoor environments. These are split into a training set of 450 scenes and a held-out test set of 50 scenes, with no scene overlap between splits. Each sample represents a complete task instance paired with a natural language instruction, recorded under varied spatial configurations and viewing conditions.

To overcome the limited scale of real-world demonstrations, we supplement the training data with 50,000 pseudo-labeled synthetic samples generated from the Rexverse-2M grounding dataset (see Section~\ref{sec:method} for details). Critically, this synthetic data is used exclusively for the SFT pre-training phase. The RL post-training stage operates solely on the 450 real-world demonstrations, using task-completion rewards to correct residual biases and improve real-world robustness.

\noindent \textbf{Baseline and Comparison Scope.}
Language-guided PTZ camera control is a task introduced by this work; to our knowledge, no existing model directly addresses this problem. Existing VLA models (e.g., OpenVLA~\cite{kim2024openvla}) target robotic manipulation and lack PTZ control interfaces, making direct comparison infeasible. Active visual search methods (e.g., Active-O3~\cite{zhu2025active}) operate purely in software by cropping regions of a static image, and thus cannot interface with physical PTZ actuators or produce real-world camera control signals.

We therefore evaluate against two categories of comparisons: (1)~a \textit{traditional ML baseline} (Random Forest) that predicts camera actions from geometric features of the target bounding box, representing a non-learned, task-specific heuristic; and (2)~\textit{ablated variants} of EyeVLA that isolate the contribution of each design choice (IoU filtering, bbox CoT guidance, and RL post-training). Together, these comparisons allow us to assess both the necessity of our learned approach and the impact of each component.

\noindent \textbf{Metrics.}
We evaluate performance along two complementary dimensions.

\textbf{(A) Mean Absolute Error (MAE).}
We report the MAE of the three predicted action dimensions:
\begin{equation}
    \text{MAE} = \frac{1}{n} \sum_{i=1}^n \lvert y_i - \hat{y}_i \rvert,
\end{equation}
where $y_i$ and $\hat{y}_i$ denote the ground-truth and predicted values, respectively. For pan ($\Delta\theta_1$) and tilt ($\Delta\theta_2$), errors are measured in degrees (°). For zoom ($\Delta z$), errors are measured in raw zoom units; a change of approximately 90 units corresponds to $1\times$ zoom in the real-world scene.

\textbf{(B) Task Completion Rate (CR).}
We deploy each model on the physical robotic eye system and evaluate it over five independent runs on the 50 held-out real-world scenes. A trial is considered successful if the target object occupies a clearly recognizable and centered region of the resulting frame, as judged by human evaluation. We report the mean task completion rate averaged over the five runs.

\noindent \textbf{Action Token Vocabulary.} 
Following the hierarchical encoding scheme described in Section~\ref{sec:method}, we extend the Qwen2.5-VL tokenizer with 43 new action tokens. Specifically, we introduce separate token sets for positive and negative displacements of $\Delta\theta_1$ and $\Delta\theta_2$ (since pan and tilt can rotate in both directions), and a positive-only token set for $\Delta z$ (as zoom only increases). 

Each action dimension is encoded using a hierarchical digit-wise decomposition based on the basis set $\{5, 2, 1\}$. For $\Delta z$, we include three digit levels (hundreds, tens, and units) to cover a wider dynamic range of zoom adjustments. For $\Delta\theta_1$ and $\Delta\theta_2$, we extend the representation to include tens, units, and one decimal place to capture fine-grained rotational control. 

This design yields a compact yet expressive vocabulary that supports both coarse and fine-grained real-world actions. The corresponding token embeddings are randomly initialized and jointly optimized during SFT.
\begin{table}[t]
\centering
\caption{Comparison of different approaches on the PTZ camera control task. ``Qwen2.5-VL (Direct)'' directly prompts the VLM to predict PTZ actions without grounding. ``Qwen2.5-VL (Grounding)'' first performs object grounding and then applies a hard-coded geometric mapping to generate actions. Lower MAE is better for $\theta_1$, $\theta_2$, and Zoom, while higher completion rate (CR) is better.}
\label{tab:qwen_compare}
\renewcommand{\arraystretch}{1.15}
\setlength{\tabcolsep}{2.5mm}
\footnotesize
\begin{tabular*}{\linewidth}{l @{\extracolsep{\fill}} cccc}
\noalign{\hrule height 1.2pt}
Model & $\theta_1$ (°) & $\theta_2$ (°) & Zoom & CR \\
\hline
Qwen2.5-VL (Direct) & -- & -- & -- & 12\% \\
Qwen2.5-VL (Grounding) & 3.87 & 2.34 & 109.46 & 70\% \\
EyeVLA (RL3) & 2.04 & 1.68 & 65.37 & 96\% \\
\noalign{\hrule height 1.2pt}
\end{tabular*}
\vspace{-1mm}
\end{table}

\subsection{Main Results}
\noindent \textbf{Quantitative Comparison.}
We first conduct a quantitative comparison of the results from three rounds of SFT, as shown in Figure~\ref{fig:3}. Specifically, we compute the MAE of the models after each SFT round on the test set with respect to the three variables: $\Delta\theta_1$, $\Delta\theta_2$, and $\Delta z$. In Table~\ref{tab:tab1}, we present a comprehensive comparison of MAE scores and task completion rates across all methods and ablation variants. We additionally compare our final model with two Qwen2.5-VL baselines in Table~\ref{tab:qwen_compare}: directly prompting the VLM to predict PTZ actions (Direct), and using its grounding capability followed by a hard-coded geometric mapping (Grounding).

As shown in Table~\ref{tab:tab1}, EyeVLA (RL3) achieves the lowest MAE across all three action dimensions ($\theta_1$: 2.04°, $\theta_2$: 1.68°, Zoom: 65.37) and the highest average task completion rate (96\%), substantially outperforming both the ML baseline (CR: 36\%) and the SFT-only variants. Table~\ref{tab:qwen_compare} reports the performance of using Qwen2.5-VL on this task. Directly prompting the VLM to predict PTZ actions without any grounding yields only 12\% CR, indicating that off-the-shelf VLMs lack the ability to reason about physical camera control. When augmented with grounding followed by a hard-coded geometric mapping, the completion rate rises to 70\%, but still falls significantly short of EyeVLA (96\%), demonstrating the advantage of our end-to-end learned approach. Notably, even SFT1 already surpasses the ML baseline in terms of MAE on $\theta_2$ and Zoom, demonstrating the strong generalization capacity of the VLM backbone. The progressive improvement from SFT1 through RL3 validates the effectiveness of each stage in the training pipeline.

We further evaluate all models on the 50 held-out real-world scenes using the physical robotic eye system over five independent runs, reporting mean task completion rates. We note that the ML baseline is provided with ground-truth bounding boxes as input features (an advantage not available to EyeVLA), yet it still achieves only 36\% CR, highlighting the necessity of our learned approach.

In Figure~\ref{fig:4}, we present visualizations of the bbox predictions and the camera-captured images after executing actions in two different scenes with different targets, corresponding to the following two questions: ``What is the brand of the tissue box?'' (requiring low-resolution information from the initial state image) and ``What is the brand of the pen inside the box?'' (where the box space is too small to approach by driving the camera and its actuators). Clearly, through the interactive iteration between the data and the model, the model's performance gradually improves.

\textbf{Analysis of Action Encoding.}
We also experimented with treating action prediction as direct continuous regression, training the model to output raw $(\Delta\theta_1, \Delta\theta_2, \Delta z)$ values without discretization. The resulting model severely overfit to the training distribution and achieved near-zero task completion on the held-out test set, confirming that the hierarchical discrete token representation is essential for generalization.

\begin{figure}[t]
\includegraphics[width=\columnwidth]{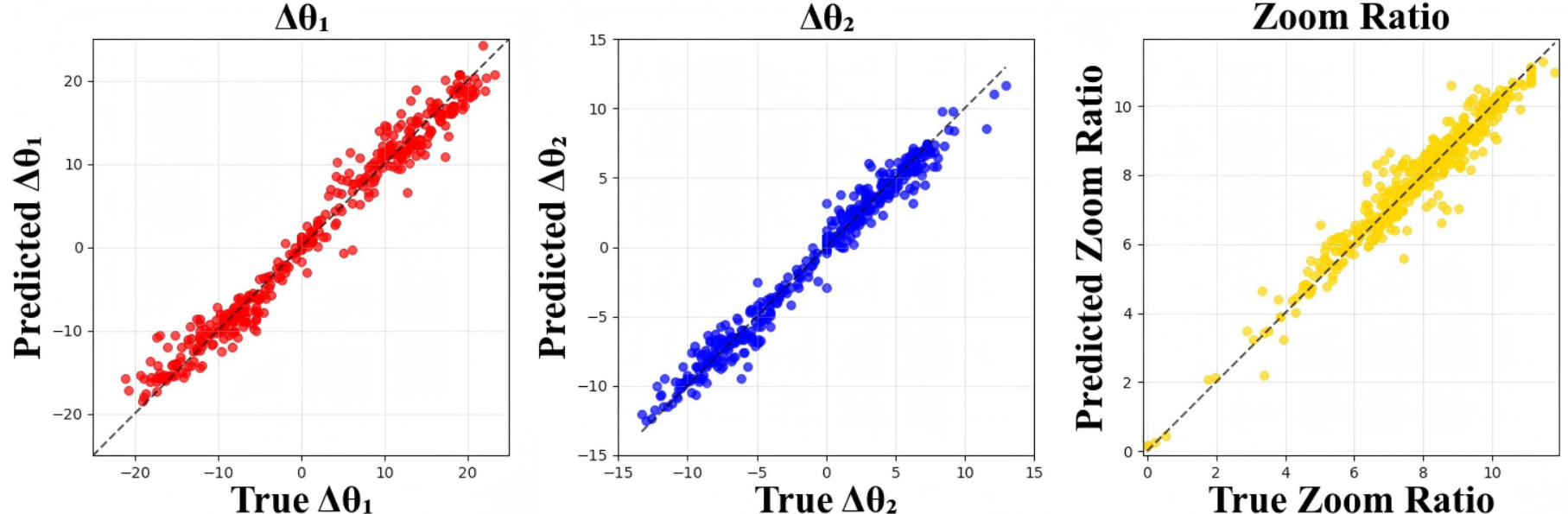}
\caption{The fitting performance of the Random Forest}
\label{fig:6}
\end{figure}
\subsection{Implementation Details}

\noindent\textbf{Pseudo-data Source.} We randomly sampled 50,000 data pairs from the Rexverse-2M dataset, an open-world object detection dataset, and filtered for instances containing a single usable target. Each pair is composed of a referring expression and a corresponding image with a ground-truth bounding box, which we use as the instruction--image input. The Random Forest model trained on our 300 most calibrated real-world demonstrations is then applied to generate pseudo-labels for $\Delta\theta_1$, $\Delta\theta_2$, and $\Delta z$.
Figure~\ref{fig:6} shows the fitted relationship, which exhibits strong linear correspondence and is well-suited for pseudo-label generation.

\noindent\textbf{Reward Function.} The total reward is the average of four sub-rewards:
\begin{equation}
R = \frac{1}{4} \left( R_{\mathrm{IoU}} + R_{\theta_1} + R_{\theta_2} + R_{\mathrm{zoom}} \right),
\end{equation}
where $R_{\mathrm{IoU}}$ is the Intersection-over-Union between the predicted and ground-truth bounding boxes (encouraging accurate spatial reasoning), and $R_{\theta_1}$, $R_{\theta_2}$, $R_{\mathrm{zoom}}$ penalize the absolute errors of the three action dimensions. For pan and tilt, the per-dimension reward increases linearly from 0 to 1 as the angular error decreases from a tolerance threshold of 1° to 0°, and decreases linearly to 0 for errors beyond 1°. For zoom, the reward is defined analogously over a wider tolerance window, reflecting the coarser precision requirements of the zoom axis.

\begin{figure}[ht]
\includegraphics[width=\columnwidth]{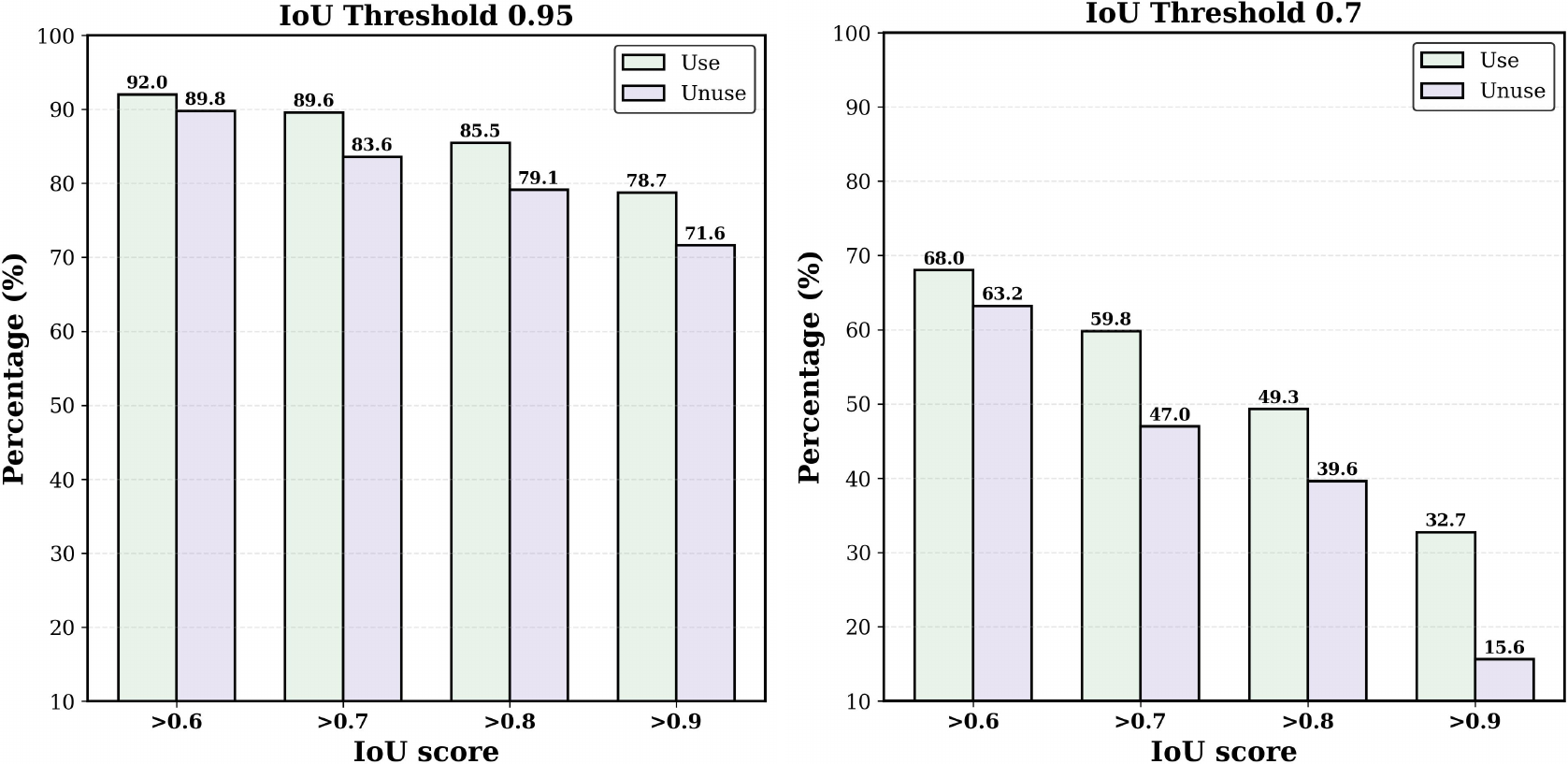}
\caption{Comparison of IoU scores between inference results with and without the replacement strategy during training, evaluated against ground-truth annotations. }
\label{fig:iou}
\end{figure}

\subsection{Ablation Study}
\noindent\textbf{IoU-Controlled Data Filtering.} To validate the effectiveness of IoU-based data filtering and bbox CoT guidance, we conduct ablation experiments by independently removing each component. As reported in Table~\ref{tab:tab1}, removing IoU filtering (SFT2(N) vs. SFT2(Y)) leads to a sharp degradation across all metrics: $\theta_1$ MAE rises from 3.97° to 4.91° and CR drops from 72\% to 56\%, indicating that low-quality pseudo-labels without IoU filtering introduce noise that harms training. The same pattern holds at the SFT3 stage. These results confirm that IoU serves as an effective proxy for pseudo-label quality, and filtering out samples with low spatial alignment is critical for iterative refinement to succeed (see also Figure~\ref{fig:iou}).

\noindent\textbf{Replacement of the bbox.} To enhance the model’s ability to accurately perceive the target location, we replace the bbox annotations in the IoU-filtered training data with ground-truth bounding boxes during data iteration, thereby strengthening the supervisory signal. We compare the model performance before and after this replacement by analyzing the resulting IoU distributions and MAE under both settings: with and without bbox replacement. The results clearly demonstrate that replacing predicted bboxes with ground-truth ones provides a more effective and positive guidance for the model.

\noindent\textbf{Reinforcement Learning.} To validate the contribution of the RL post-training stage, we compare SFT3 and RL3 in Table~\ref{tab:tab1}. Despite SFT3 already achieving competitive MAE scores, RL3 further reduces $\theta_1$ error from 2.94° to 2.04° and improves the average CR from 92\% to 96\%. As shown in Table~\ref{tab:table3}, RL consistently improves the mean IoU of predicted bounding boxes at both stages (SFT2$\to$RL2: 0.68$\to$0.75; SFT3$\to$RL3: 0.91$\to$0.93). We hypothesize that SFT on pseudo-labels causes the model to overfit to the label distribution, occasionally collapsing action predictions to a narrow range. RL with real-world reward signals breaks this pattern by penalizing systematic biases, leading to more diverse and physically grounded action outputs. This effect is especially pronounced for zoom, where the reward tolerance window encourages the model to explore a broader range of zoom adjustments.
\begin{table}[t]
\centering
\renewcommand{\arraystretch}{1.5}
\setlength{\tabcolsep}{2.5mm}
\caption{Comparison of the average IoU between the two stages with and without RL.}
\begin{tabular}{lcccc}
\toprule
   & SFT2  & RL2  & SFT3  & RL3\\
\midrule
IoU & 0.68 & 0.75 & 0.91 & 0.93\\
\bottomrule
\end{tabular}
\label{tab:table3}
\end{table}

\section{Conclusion}

In this work, we propose EyeVLA, a novel language-guided active vision system for embodied perception. This framework realizes open-world perception and understanding specified by natural language instructions through joint decision-making between the VLA model and hardware actuators. It enables intelligent agents to actively acquire more informative observational data under constraints of pixel and spatial budgets, effectively addressing a key limitation in current vision systems: their passive reliance on fixed inputs and inability to adaptively gather optimal visual information in dynamic, complex environments.

We design an efficient hierarchical action encoding-decoding scheme and an iterative pipeline based on hybrid training and data filtering, which enables the generation of large amounts of realistic augmented data from a small amount of real-world data, thereby ensuring high semantic quality and supervisability of the data. Experiments on 50 diverse real-world scenes demonstrate the effectiveness and generalization of our method. We hope EyeVLA can serve as a practical approach and a new direction for instruction-driven active visual perception in embodied AI.

\begin{acks}
Acknowledgments are omitted for anonymous review.
\end{acks}

\bibliographystyle{ACM-Reference-Format}
\bibliography{main}

\end{document}